\documentclass{datadogtech}

\usepackage{amsmath}
\usepackage{amssymb}
\usepackage{amsthm}
\usepackage{mathtools}
\usepackage{xspace}

\newcommand{\toto}{Toto\xspace}
\newcommand{\totoone}{Toto~1.0\xspace}
\newcommand{\totov}{Toto~2.0\xspace}

\newcommand{\ff}{Toto 2.0 FnF\xspace}

\newcommand{\boom}{\textsc{BOOM}\xspace}
\newcommand{\gifteval}{GIFT-Eval\xspace}

\newcommand{\ump}{u-$\mu$P\xspace}
\newcommand{\mup}{$\mu$P\xspace}
\newcommand{\normuon}{NorMuon\xspace}
\newcommand{\adamw}{AdamW\xspace}

\newcommand{\cmax}{\ensuremath{c_{\max}}}
\newcommand{\pmax}{\ensuremath{p_{\max}}}
\newcommand{\dmodel}{\ensuremath{d_{\text{model}}}}
\newcommand{\dk}{\ensuremath{d_k}}
\newcommand{\dhead}{\ensuremath{d_{\text{head}}}}
\newcommand{\nheads}{\ensuremath{h}}
\newcommand{\nlayers}{\ensuremath{L}}

\title{Toto~2.0: Time Series Forecasting Enters the Scaling Era}

\author[*,\dagger,1]{Emaad Khwaja}
\author[*,1]{Chris Lettieri}
\author[*,\dagger,1]{Gerald Woo}
\author[1]{Eden Belouadah}
\author[1]{Marc Cenac}
\author[\ddag]{Guillaume Jarry}
\author[\ddag]{Enguerrand Paquin}
\author[1]{Xunyi Zhao}
\author[1]{Viktoriya Zhukova}
\author[1]{Othmane Abou-Amal}
\author[1]{Chenghao Liu}
\author[1,2]{Ameet Talwalkar}
\author[1]{David Asker}

\affiliation[1]{Datadog AI Research}
\affiliation[2]{Carnegie Mellon University}

\contribution[*]{Core Contributor, listed alphabetically}
\contribution[\dagger]{Correspondence: \href{mailto:emaad@datadoghq.com,gerald.woo@datadoghq.com}{\texttt{\{emaad,\,gerald.woo\}@datadoghq.com}}}
\contribution[\ddag]{Work completed during internship at Datadog}

\abstract{%
We show that time series foundation models \textit{scale}: a single training recipe produces reliable forecast-quality improvements from 4m to 2.5B parameters. We release \totov{}, a family of five open-weights forecasting models trained under this recipe.  The \totov{} family sets a new state of the art on three forecasting benchmarks: \boom{}, our observability benchmark; \gifteval{}, the standard general-purpose benchmark; and the recent contamination-resistant TIME benchmark.
This report describes our experimental results and details the design decisions behind \totov{}: its architecture and training recipe, training data, and the \ump{} hyperparameter transfer pipeline. All five base checkpoints are released under Apache~2.0.
}

\date{{\sffamily\today}}
\metadata[Code]{\href{https://www.github.com/DataDog/toto}{\nolinkurl{https://www.github.com/DataDog/toto}}}
\metadata[Weights]{\href{https://www.huggingface.co/collections/Datadog/toto-20}{\nolinkurl{https://www.huggingface.co/collections/Datadog/toto-20}}}

\begin{document}

\maketitle

\begin{figure}[H]
    \centering
    \includegraphics[width=\linewidth]{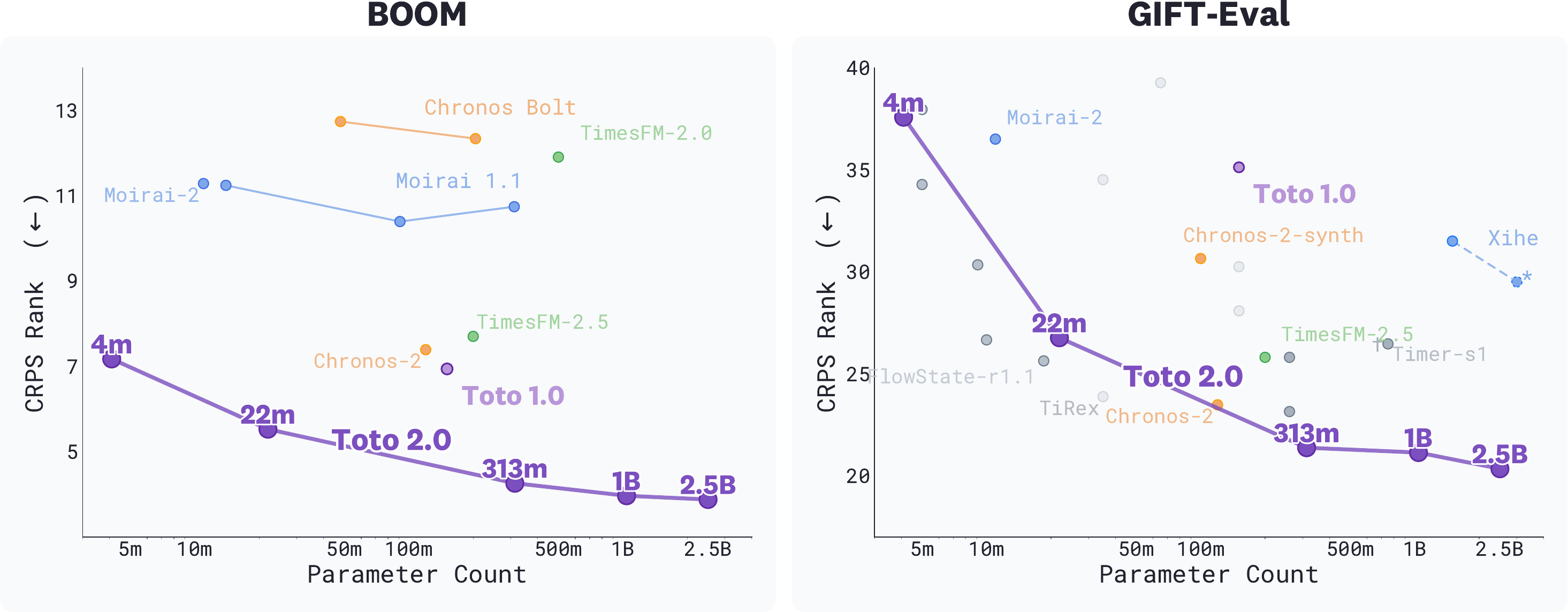}
    \caption{CRPS rank vs.\ parameter count on \boom{} (\textbf{left}) and \gifteval{} (\textbf{right}) for top foundation models; lower is better. \totov{} is the only family whose performance improves reliably with scale, with every size sitting on or near the Pareto frontier of both benchmarks. Competing model families scale unevenly, with larger versions sometimes underperforming smaller ones. \\ [0.4em] {\footnotesize $^{*}$Xihe-ultra parameter count estimated ($\sim$3B); not officially disclosed. $^{\dagger}$Timer-s1 is an 8.3B mixture-of-experts model (750m active).}}
    \label{fig:pareto-combined}
\end{figure}

\section{Introduction}
\label{sec:introduction}

Over the past year, time series foundation models (TSFMs) have begun to match or exceed tuned statistical baselines across heterogeneous domains, much as BERT~\citep{devlin2019bert} did for language a decade ago~\citep{berts2025workshop}. What TSFMs have \emph{not} yet replicated from NLP and vision is reliable scaling: a single recipe applied at successively larger widths and token budgets that produces predictable returns~\citep{radford2019gpt2,kaplan2020scaling}.

We present \totov, a family of five open-weights forecasting models (4m, 22m, 313m, 1B, and 2.5B parameters) designed to answer a simple, open question: can TSFMs improve from scaling? Our results show they do. Every size improves on the one below it (\Cref{fig:pareto-combined}). \totov{} takes the top spots on every benchmark we evaluated: \boom~\citep{cohen2025this}, \gifteval~\citep{aksu2024gifteval}, and TIME~\citep{qiao2026time}. The family is also a generational jump from \totoone{}: the 22m matches \totoone{}'s quality with $7\times$ fewer parameters, and inference is dramatically faster at long horizons. \totov{} sees no public forecasting data during pretraining. It trains exclusively on Datadog observability metrics and synthetic series, yet leads the field on general-purpose benchmarks.

The remainder of this report is organized as follows.
\begin{itemize}
    \item \textbf{Architecture and training recipe} (\Cref{sec:method}). \totov{} refines the \totoone{} backbone in three key aspects: contiguous patch masking (CPM) replaces autoregressive decoding to enable single-pass parallel forecasting; a quantile output head replaces the Student-T mixture of \totoone{} to improve stability at scale; and \normuon{} replaces \adamw{} to better match the new loss function (\eqref{eq:pinball}) used for fitting the quantile head.
    \item \textbf{Training data} (\Cref{sec:data}). Unlike other leading TSFMs, we do not pretrain on any public time series data, and instead rely exclusively on a mix of Datadog's internal observability metrics and synthetic data. Public data enters the recipe only during finetuning, where it makes up 45\% of the mix (\Cref{sec:gifteval_ft}). This makes \totov{}'s public-benchmark performance a stronger test of cross-domain generalization than for models pretrained directly on public time-series corpora: the base models have never seen any public evaluation domains, yet generalize to them.
    \item \textbf{Hyperparameter transfer pipeline} (\Cref{sec:hp_transfer}). We built a structured search procedure that tunes hyperparameters once on a 10m proxy and transfers the same configuration to all five target sizes, modifying width, depth, and head count. The transfer is enabled by \ump{}, which makes learning dynamics width-independent.
    \item \textbf{Results and scaling behavior} (\Cref{sec:results}). \totov{} sets a new state of the art on \boom{}, \gifteval{}, and TIME, with every size on or near the Pareto frontier. Finetuned and ensembled variants additionally top the full \gifteval{} leaderboard outright. Inference is dramatically faster than \totoone{} at long horizons, and we show larger models notably produce coherent forecasts well past their training context on synthetic multi-scale signals.
    \item \textbf{Where TSFMs go next} (\Cref{sec:discussion}). We share our view of the next set of bottlenecks and opportunities: closing the long-horizon gap with classical baselines, data curation, evaluation that tracks downstream value, and multimodality.
\end{itemize}

\paragraph{Releases.} Model weights for all five sizes are available at \href{https://huggingface.co/collections/Datadog/toto-20}{https://huggingface.co/collections/Datadog/toto-20}, and our distributed training library is released at \href{https://github.com/DataDog/toto/tree/main/dd_unit_scaling}{\texttt{dd\_unit\_scaling}} under Apache~2.0.

\section{Architecture}
\label{sec:method}

\begin{figure}[t]
    \centering
    \includegraphics[width=\linewidth]{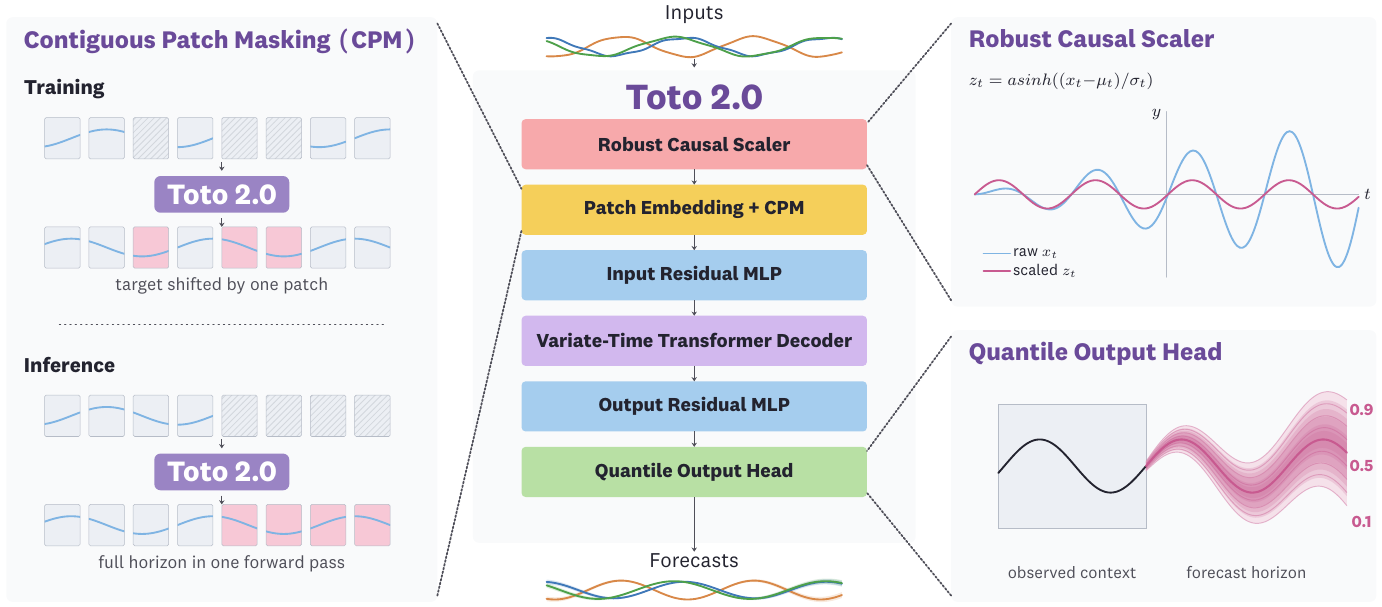}
    \caption{\totov{} architecture. \textbf{Left:} \emph{training and inference protocol}: CPM training applies variable-length contiguous masked spans to the input; at inference the horizon is filled with mask tokens and decoded in a single forward pass.
    \textbf{Center:} \emph{forward pass}: a decoder-only transformer with alternating time-axis (causal) and variate-axis (full) attention, retained from \totoone{}. The input scaler, patch projections, masking strategy, and output head are all improvements on the \totoone{} backbone.
    \textbf{Right:} \emph{input and output heads}: a robust causal scaler ($\operatorname{arcsinh}$ normalization) on the input side, and a quantile output head producing nine quantile levels.}
    \label{fig:architecture}
\end{figure}

The \totov{} backbone is largely retained from \totoone{}~\citep{cohen2024toto}: a decoder-only patched transformer whose attention layers alternate between time-axis (causal) and variate-axis (full) views of the input. The main changes include: contiguous patch masking for parallel decoding (\Cref{sec:cpm}), a quantile output head replacing the Student-T mixture (\Cref{sec:quantile_head}), \normuon{} replacing \adamw{} (\Cref{sec:optimizer}), amongst others (\Cref{sec:misc}).

\subsection{Contiguous patch masking}
\label{sec:cpm}

\totov{} (\Cref{fig:architecture}) replaces \totoone{}'s autoregressive decoding with \emph{contiguous patch masking}, an elegant single-pass parallel scheme adapted from \citet{auer2025tirex}. In \totoone{}, the model $f_\theta$ extends $N$ context patches $\mathbf{p}_{1:N}$ of size $P$ one patch at a time via $\hat{\mathbf{p}}_i = f_\theta(\mathbf{p}_{1:i-1})$. A $H$-step horizon takes $K = H/P$ sequential calls, which is both slow and fragile to errors compounding across the $K$ steps. CPM addresses both: train with variable-length masked spans so the model learns to predict multiple future patches at once. Each patch carries a binary mask channel $\mathbf{b}_i \in \{0,1\}^P$ with $b_{i,k} = 1$ at unobserved entries and $0$ elsewhere. For CPM-masked positions $\mathcal{M} \subseteq \{1, \ldots, N\}$:
\begin{equation}
    \hat{\mathbf{p}}_i \;=\; \bigl[f_\theta(\mathbf{p}_{1:N},\, \mathbf{b}_{1:N})\bigr]_i, \qquad i \in \mathcal{M},
    \label{eq:cpm}
\end{equation}
with the loss (\Cref{eq:quantile_loss}) averaged over all $N$ positions. CPM pays off more with a transformer than on the xLSTM~\citep{beck2024xlstm} it was designed for: \Cref{eq:cpm} is one call to $f_\theta$ with a transformer, $|\mathcal{M}|$ on a SSM. At train time, $\mathcal{M}$ is sampled as random contiguous spans length $c \sim \mathcal{U}\{1{:}\cmax\}$ with probability $p \sim \mathcal{U}(0, \pmax)$. At inference, $\mathcal{M} = \{N+1, \ldots, N+K\}$. Either way, the model commits to a coherent forecast all at once, mitigating the compounding error of autoregressive decoding.

For horizon lengths where single-pass decoding may lose coherence, \totov{} also supports \emph{block decoding}: apply \Cref{eq:cpm} round by round in blocks of $B$ patches, committing $\mathbf{p}_i \leftarrow \mathrm{median}(\hat{\mathbf{p}}_i)$ and $\mathbf{b}_i \leftarrow 0$ for $i \in \mathcal{M}$ after each round (KV cache is reused). This incurs $B-1$ more forward passes but mitigates overall drift. We find single-pass generally remains stable up to a $\sim$768-step horizon (on synthetic multi-scale signals). We use block decoding for the long-horizon study in \Cref{sec:long_horizon}.

Our sweeps (\Cref{sec:hp_transfer}) found optimal settings of $\cmax{}=16$ and $\pmax{}=0.4$, versus TiRex's $\cmax{}=5$ and $\pmax{}=0.25$, suggesting \totov{} can handle longer masked spans than the recurrent schema TiRex was originally designed with.

\subsection{Quantile output head}
\label{sec:quantile_head}

\totoone{} used a Student-T mixture model (SMM) to produce probabilistic forecasts. The SMM worked well at the size of \totoone{}, but as we scaled beyond the original recipe, we encountered practical limits: the SMM becomes numerically unstable at large activations and diverges when predictions approach zero due to the variance term in its normalization. These issues surfaced during training as we pushed toward larger models and broader data mixes.

\totov{} replaces SMM with a quantile output head: for each future timestep, the model predicts nine quantile levels at $\mathcal{T} = \{0.1, 0.2, \ldots, 0.9\}$, trained with the pinball loss~\citep{koenker1978regression}. For a target value $y$ and predicted quantile $\hat{q}_\tau$, the pinball loss at level $\tau$ is
\begin{equation}
    \rho_\tau(y - \hat{q}_\tau) \;=\; (y - \hat{q}_\tau)\bigl(\tau - \mathbb{1}[y < \hat{q}_\tau]\bigr),
    \label{eq:pinball}
\end{equation}
and the head loss averages over the nine levels:
\begin{equation}
    \mathcal{L}_{\text{quantile}} \;=\; \frac{1}{|\mathcal{T}|} \sum_{\tau \in \mathcal{T}} \rho_\tau(y - \hat{q}_\tau).
    \label{eq:quantile_loss}
\end{equation}
Quantile heads are now standard among leading TSFMs~\citep{ansari2025chronos2,google2025timesfm,liu2025moirai2} for their stability and calibration. We sort the predicted quantiles during inference to prevent crossing.

\subsection{Optimizer}
\label{sec:optimizer}
\totov{} uses \normuon{}~\citep{li2025normuon} to optimize all matrix-shaped parameters. We argue this choice particularly well-suited to pinball training; the rest of this section develops the reasoning.
 
\totoone{} trained with \adamw~\citep{loshchilov2019adamw} on the negative log-likelihood (NLL) of its SMM. The pairing was natural: NLL provides smooth, magnitude-bearing gradients, and \adamw{} is the default optimizer for nearly all foundation models. With \totov{}'s switch to pinball, that pairing becomes less effective: pinball's sign-valued gradients narrow the dynamic range over which \adamw{}'s variance-driven step-size mechanism operates. Differentiating \Cref{eq:pinball} gives
\begin{equation}
    \frac{\partial \rho_\tau(y - \hat{q})}{\partial \hat{q}} = g_\tau \;=\; \begin{cases} -\tau & y > \hat{q}, \\ 0 & y = \hat{q}, \\ 1 - \tau & y < \hat{q}, \end{cases}
    \label{eq:pinball_grad}
\end{equation}
which takes only three values regardless of $|y - \hat{q}|$. Contrast this with the MSE gradient, $\frac{\partial (y - \hat{q})^2}{\partial \hat{q}} = -2(y - \hat{q})$, whose magnitude scales linearly with the error. Two residuals differing by an order of magnitude produce gradients differing by an order of magnitude under MSE, but identical-magnitude gradients under pinball. With sign-valued gradients, the loss provides a direction to refine the model towards, but not how wrong it is, so the optimizer has to infer step size from its own internal states.
 
One possible explanation for AdamW's weaker performance in this setting comes from \citet{balles2020dissecting}, who decompose Adam~\citep{kingma2017adam} into two aspects: ``for each weight, the update direction is determined by the sign of stochastic gradients, whereas the update magnitude is determined by an estimate of their relative variance ($v_t$).'' Under the sign-valued gradients of \Cref{eq:pinball_grad}, this is the only step-size mechanism Adam has: the per-step gradient carries no magnitude information, so all per-weight scale adaptation comes from $v_t$. Adam trains successfully in this regime, but with limited dynamic range.
 
Muon~\citep{jordan2024muon} has emerged as the leading post-\adamw{} candidate for large-scale training, with roughly $2\times$ compute-efficiency gains over \adamw{} in scaling-law experiments and adoption at trillion-parameter scale by Moonshot AI's Kimi K2~\citep{liu2025muonscalable}. For a 2D weight $W$ with matrix gradient $G_t$, Muon maintains a momentum buffer $B_t = \mu B_{t-1} + G_t$, orthogonalizes it via a Newton--Schulz iteration $O_t = \mathrm{NS}(B_t)$ that drives the singular values of $B_t$ toward unity, and applies $W_t \;\leftarrow\; W_{t-1} - \eta\, O_t$.

Muon contains no second-moment EMA, discarding Adam's $\beta_2$ variance mechanism by design. On smooth losses, this is part of what gives Muon its compute-efficiency advantage over \adamw{}, and is part of why the broader community has adopted it. In our pinball-loss setting, this tradeoff appears less favorable: removing the variance mechanism entirely also removes the limited step-size adaptation that remained.
 
Although Newton--Schulz drives the singular values of $B_t$ toward unity, the per-row $L^2$ norms of $O_t$ can still vary by orders of magnitude, so a handful of neurons dominate each update. \normuon{}\footnote{\normuon{} has also been gaining traction more broadly: Andrej Karpathy's \href{https://github.com/karpathy/nanochat/discussions/481}{\texttt{nanochat}} uses it to train GPT-2 for under \$100~\citep{karpathy2026nanochat}.} balances per-neuron contributions by normalizing each row of $O_t$ against an EMA of its own squared magnitude:
\begin{equation}
\begin{aligned}
    v_t \;&=\; \beta_2\, v_{t-1} + (1 - \beta_2) \cdot \mathrm{mean\_cols}(O_t \odot O_t), \\
    W_t \;&\leftarrow\; W_{t-1} - \eta \, O_t \big/ \sqrt{v_t + \epsilon},
\end{aligned}
\label{eq:normuon}
\end{equation}
where $\odot$ denotes the Hadamard product, $\mathrm{mean\_cols}$ reduces each row of $O_t \odot O_t$ to its column-mean (yielding a per-row scalar), and the division and square root in the update are applied row-wise via broadcasting. \normuon{}'s row normalization, motivated by per-neuron balancing, also reinstates the $\beta_2$ variance mechanism---now applied per neuron rather than per parameter. This contrasts with Adam, whose parameter-wise $v_t$ never leaves the single weight it indexes and has no view of how weights within a neuron relate to each other.
 
We use \normuon{} for all internal matrix-shaped parameters and \adamw{} for input/output projections, biases, and norms. We use Nesterov momentum and replace the standard Newton--Schulz orthogonalization with Polar Express~\citep{amsel2026polar}, a quintic iteration with coefficients optimized for faster convergence of the singular values to unity at low precision. Following $\mu$P++~\citep{ren2025muppp}, we do not apply weight decay to biases, norms, or input/output projection weights. For other parameters, we apply cautious weight decay~\citep{chen2025cwd}, which applies decay only to parameters whose signs align with the optimizer update.

\subsection{Additional architectural changes}
\label{sec:misc}

Four more changes round out the redesign:
\paragraph{Patch size.} \totov{} uses a patch size of 32, down from 64 in \totoone{}. This doubles the sequence length the transformer sees for a given input window, allowing the model to learn finer-grained representations of within-patch dynamics at the cost of longer attention computations.

\paragraph{Robust input normalization.} Observability metrics routinely span many orders of magnitude. Request rates can move from tens to millions per second, latencies from microseconds to seconds. \totoone{} handled this with a novel causal normalization mechanism. \totov{} enhances this by adding a robust $\operatorname{arcsinh}(z) = \log\!\bigl(z + \sqrt{z^2 + 1}\bigr)$ transformation ~\citep{ansari2025chronos2}, which behaves as $z$ for $|z| \ll 1$ and as $\operatorname{sign}(z)\log(2|z|)$ for $|z| \gg 1$. The model predicts in this scaled space, and predictions are unscaled to compute the final forecast. Small fluctuations near zero are thus preserved at full resolution while large excursions are compressed logarithmically, all without discarding sign information.

\paragraph{Residual MLP patch projections.} \totoone{} used linear layers for both patch embedding (mapping raw patches to model-dimension vectors) and output projection (mapping model-dimension vectors to distribution parameters). \totov{} replaces both with two-layer SiLU networks with residual connections, giving the model nonlinear patch representations at both ends of the transformer.

\paragraph{Attention changes.} We add PerDimScale (learned per-dimension query scaling, also used in TimesFM~2.5 \citep{google2025timesfm}) with $1/\dk$ attention scaling for \mup~\citep{yang2021tensor} compatibility. Patches with entirely missing observations are masked out of attention computation. Bias terms are enabled on attention projections but not on MLPs, and dropout is not used during training.

\section{Training data}
\label{sec:data}

\begin{figure}[t]
    \centering
    \includegraphics[width=1\linewidth]{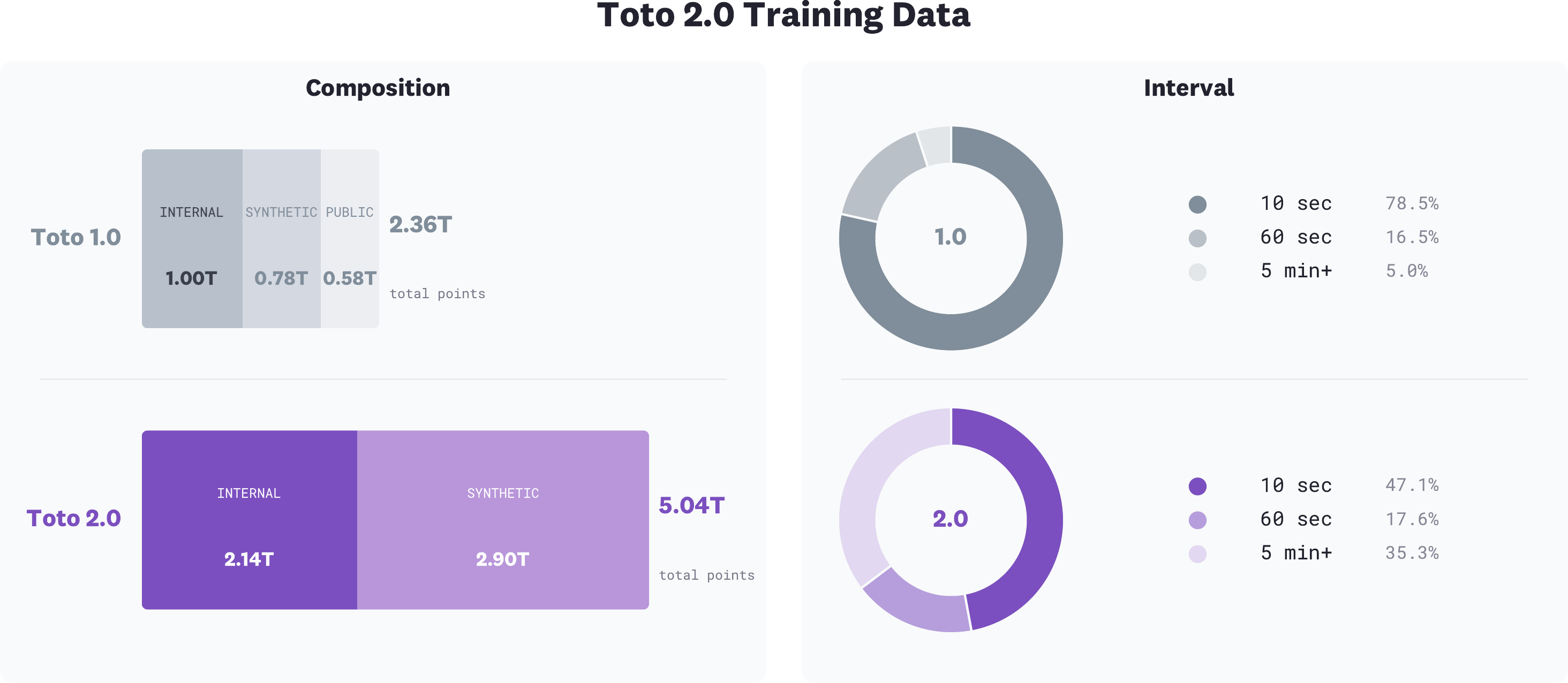}
    \caption{Training data composition for \totoone{} (2.36\,T points) and \totov{} (5.04\,T points for the 313m, 1B, and 2.5B; 3.40\,T points for the 4m and 22m). \textbf{Left:} \totov{} composition shown is for the 5.04\,T mix used by the three largest models; the 3.40\,T mix used by the 4m and 22m holds the relative proportions constant. \totov{} drops public data entirely; internal observability metrics roughly double, and synthetic data nearly quadruples compared to \totoone{}. \textbf{Right:} sampling-interval breakdown of the internal observability portion only (2.14\,T points of \totov{}, vs.\ the corresponding \totoone{} subset); percentages are within this subset rather than the full training mix. \totov{} rebalances away from high-frequency intervals: 5\,m+ data rises from 5\% to 35\%, while 10\,s data drops from 78.5\% to 47.1\%.}
    \label{fig:training_data}
\end{figure}

\totov{} trains exclusively on a mix of Datadog's internal telemetry and synthetic data. Our larger models (313m, 1B, 2.5B) see 5.04\,T data points and our smaller ones (4m, 22m) see 3.40\,T, up from 2.36\,T in \totoone{} (\Cref{fig:training_data}).

We made two structural changes from \totoone{}. First, we removed all public data from pretraining. Our hyperparameter sweep (\Cref{sec:hp_round2}) found that public time series data was suboptimal at proxy model scale; the best mixtures the sweep found excluded it entirely. Public data does, however, enter the finetuning recipe of Toto 2.0 2.5B-FT, where is makes up 45\% of the mix (\Cref{sec:gifteval_ft}). Second, we more than doubled our synthetic data using newer generation methods that produce more diverse regimes.

We also rebalanced the internal Datadog telemetry data. \totoone{}'s mix skewed heavily toward high-frequency (10\,s) intervals. For \totov{} we parameterized the sampling interval and overweighted longer intervals, so the model sees a more diverse, higher-signal view of the same underlying telemetry.

\subsection{Observability time series from Datadog}
\label{sec:data_obs}

\totov{}'s real-world training data comes exclusively from Datadog's own internal observability metrics: CPU utilization, memory usage, request latency, error rates, and similar infrastructure signals. Compared to \totoone{}, the dataset is larger, draws from a broader set of data sources, and covers more recent time periods. No customer data is used at any point.

\subsection{Synthetic data}
\label{sec:data_synth}

\totoone{}'s synthetic training data used generic stochastic processes similar to \citet{das2024timesfm}. \totov{} uses the synthetic data generation method from TempoPFN~\citep{moroshan2025tempopfn}, built on the prior-data fitted network (PFN) framework~\citep{muller2022pfn} in which a transformer is trained on samples drawn from a hand-crafted prior. The TempoPFN prior is rich with nonstationary trends, abrupt changepoints, and long-range dependencies. The final training mix for base models is 42.5\% observability data and 57.5\% synthetic data, with the observability portion further split across sampling intervals as detailed in \Cref{sec:hp_round2}.

\section{Hyperparameter transfer pipeline}
\label{sec:hp_transfer}

Scaling models to multiple sizes lets users trade off inference cost against forecast quality, but this is only useful if each size is reliably better than the last. Achieving this kind of scaling behavior efficiently is notoriously difficult, and for TSFMs in particular it has been a recurring gap. Critical hyperparameters such as the learning rate are not stable across model widths under standard parametrization---empirically, the optimal learning rate can shift by an order of magnitude across width sweeps~\citep{yang2021tensor}. The naive approach, tuning hyperparameters independently for each of the five target sizes, would be inefficient: each target model requires days of training, making a large hyperparameter search computationally expensive at that scale. To turn the architectural improvements into a reliable scaling recipe, we sought a way to transfer hyperparameters across widths. For that, we turned to \ump~\citep{blake2025ump}.

\begin{figure}[!t]
    \centering
    \includegraphics[width=\linewidth]{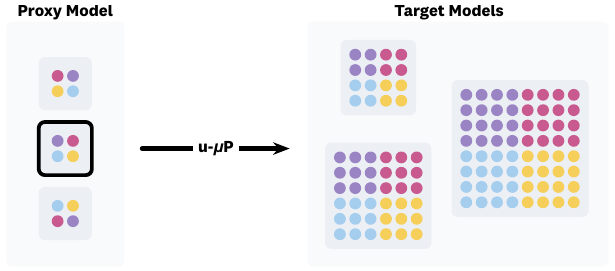}
    \caption{\ump{} makes optimal hyperparameters independent of model width. We tune parameters on a small proxy model, select the best configuration (depicted with a black outline here) and directly transfer the same configuration to any larger target model with no retuning required.}
    \label{fig:mu_transfer}
\end{figure}

\ump combines Maximal Update Parametrization (\mup)~\citep{yang2021tensor,yang2021tensorprograms4} with unit scaling~\citep{blake2023unitscaling} to make the optimal learning rate independent of model width. We selected the unit-scaled variant because of its simplicity and improved transfer for decoder-only models. \textbf{This approach allowed us to sweep hyperparameters on a cheap 10m proxy, then transfer the configuration directly to all five target sizes (\Cref{fig:mu_transfer}) in a largely automated fashion.} To our knowledge, this is the first application of \mup{} to time series forecasting.

\subsection{The proxy model}

The proxy is a 10m-parameter model ($\nlayers = 12$, $\dmodel = 256$, $\nheads = 4$). We chose a $\dmodel = 256$ because \citet{blake2025ump} demonstrates this as a floor to prevent optimal parameter drift. Each sweep trial trains the proxy for 30{,}000 steps at the same batch size used for the target models, under a warmup-stable-decay (WSD)~\citep{hu2024minicpm} learning-rate schedule. At this scale, each training run completes in a few hours rather than days, enabling a configuration search several orders of magnitude broader than would be tractable at the target sizes.

\subsection{Structured hyperparameter search}

Even at proxy scale, the joint search space spans 17 continuous and several categorical dimensions ($\sim 10^{19}$ configurations under a modest grid discretization), making exhaustive search intractable. We split the search into four sequential rounds, each one selecting the empirical optimum for a different group of decisions on top of the previous round's best configuration. The order follows the natural dependency chain: architecture and data shape the loss landscape, the optimizer must adapt to that landscape, and the decay schedule is tuned downstream of the optimized stable regime. All four rounds use Optuna~\citep{akiba2019optuna} with Tree-Structured Parzen Estimator (TPE)~\citep{watanabe2023tpe} sampling, optimizing against seasonal-naive-normalized MASE and CRPS on the \gifteval{} validation set.

\paragraph{Round 1: Architecture.}
We swept attention normalization (PerDimScale, QK-Norm~\citep{henry2020qknorm}, or neither), how often the variate-axis attention layer appears in the layer stack, which transformer layers carry bias terms, and the contiguous-patch-masking parameters. The proxy's twelve layers allowed clean exploration of several variate-attention cadences (every 2, 3, 4, 6, or 12 layers).

The best configuration uses PerDimScale (over QK-Norm), places the variate-axis attention layer last in the stack, and sets the contiguous-patch-masking parameters to $\cmax{} = 16$ and $\pmax{} = 0.4$ (longer masked spans than TiRex's defaults).

\paragraph{Round 2: Data mixture.}
\label{sec:hp_round2}
We parameterized the training mix as a constrained probability simplex over five sources, with each lower bound set to 0 so TPE could remove a source entirely if optimal:
\begin{lstlisting}[language=yaml]
sweep:
  dd_10s:      0.0 - 1.0   # Datadog 10-second metrics
  dd_60s:      0.0 - 0.7   # Datadog 60-second metrics
  dd_long:     0.0 - 0.2   # Datadog 5+ minute metrics
  synthetic:   0.0 - 1.0   # TempoPFN data
  public:      0.0 - 0.05  # GIFT-Eval Pretrain
constraint:  sum = 1.0
\end{lstlisting}
Upper bounds on the smaller corpora are set to cap repetition during training.

The optimal mixture excluded public data and settled at 42.5\% Datadog observability data and 57.5\% synthetic, with the Datadog portion split across 10\,s (20\%), 60\,s (7.5\%), and 5+\,m (15\%) intervals. This is the mix used for all base models.

\paragraph{Round 3: Optimizer.}
Starting from Round 2's best configuration, we swept the learning rate, weight decay, and first- and second-moment exponential decay rates ($\mu$ and $\beta_2$ for \normuon{}; $\beta_1$ and $\beta_2$ for \adamw{}), along with shared warmup steps and gradient clipping threshold.

The best configuration for \normuon{} is $\eta = 0.65$\footnote{The \normuon{} learning rate looks large at first glance, but is in the expected range under \ump{}: unit scaling absorbs the $1/\sqrt{\texttt{fan\_in}}$ factor into the parametrization itself, so the user-facing $\eta$ is the per-tensor update size at unit scale rather than the unnormalized step that an unconstrained optimizer would take.}, $\mu = 0.96$, $\beta_2 = 0.999$, weight decay $= 2 \times 10^{-8}$, and for \adamw{} is $\eta = 0.012$, $\beta_1 = 0.91$, $\beta_2 = 0.972$. Warmup is 6{,}000 steps with gradient clipping at 7.0.

\paragraph{Round 4: Decay schedule.}
Starting from a checkpoint inside the stable portion of Round 3's best run, we swept the length and shape (linear vs.\ \texttt{1-sqrt}) of the learning-rate decay.

Linear decay won; the final schedule decays linearly over 10{,}500 steps---a short tail relative to the total training budget (1.7--2.6\% of the 400{,}000 and 600{,}500 total steps in \Cref{tab:model_sizes}). We maintain 10{,}500 decay steps for all base models.

\subsection{Zero-shot transfer to target sizes}

\textbf{Scaling up is straightforward:} take the proxy's best configuration and apply it to every target size. The main architectural changes between sizes are embedding dimension \dmodel, depth \nlayers, and head count \nheads{} (we fix the head dimension at $\dhead = 64$). 

Under \ump{}, each hidden weight is reparametrized as $W = A_W \cdot w$ with $w_0 \sim \mathcal{N}(0, 1)$, and updated as $w_{t+1} = w_t + C_W \cdot \Phi_t$, where $\Phi_t$ is the optimizer's step direction on the gradient history. For hidden weights, the multipliers scale as $A_W \propto 1/\sqrt{\texttt{fan\_in}}$ and $C_W \propto \eta/\sqrt{\texttt{fan\_in}}$ (see Table~2 of \citet{blake2025ump} for the input/output and depth-dependent variants), which makes the optimal learning rate $\eta$ invariant across widths. Weight decay is selected at proxy scale and held fixed; it is not guaranteed by \ump{} to transfer. 

\Cref{tab:model_sizes} lists the five resulting model configurations:

\begin{table}[H]
\centering
\small
\setlength{\tabcolsep}{8pt}
\begin{tabular}{@{}lccccc@{}}
\toprule
\textbf{Model} & \textbf{\dmodel} & \textbf{\nheads} & \textbf{\nlayers} & \textbf{Training steps} & \textbf{Norm $\varepsilon$} \\
\midrule
\textbf{4m}   & 256  & 4  & 4  & 400{,}000 & $1 \times 10^{-4}$ \\
\textbf{22m}  & 512  & 8  & 6  & 400{,}000 & $1 \times 10^{-4}$ \\
\textbf{313m} & 1024 & 16 & 24 & 600{,}500 & $1 \times 10^{-4}$ \\
\textbf{1B}   & 1536 & 24 & 36 & 600{,}500 & $5 \times 10^{-4}$ \\
\textbf{2.5B} & 2048 & 32 & 48 & 600{,}500 & $5 \times 10^{-4}$ \\
\bottomrule
\end{tabular}
\caption{\totov{} model sizes. \dmodel{} is the embedding (hidden) dimension, \nheads{} the number of attention heads, and \nlayers{} the depth (number of transformer blocks); the head dimension is fixed at $\dhead = 64$ for all sizes. All five sizes train on 4{,}096-timestep contexts with patch size 32 and 32 variates per sample, at a global batch size of 64. The 4m and 22m converged at 400{,}000 steps; the larger sizes were still improving past that point and trained for 600{,}500.}
\label{tab:model_sizes}
\end{table}

\subsection{Making u-\texorpdfstring{$\boldsymbol{\mu}$}{μ}P work in production}
\label{sec:dd_unit_scaling}

The upstream \texttt{unit\_scaling} library~\citep{graphcore2023unitscaling} used for implementing \ump{} targets single-GPU eager-mode. Training large models at scale often requires \texttt{torch.compile}, model sharding, and distributed parallelism strategies for optimal speed and memory utilization. \ump{} works by attaching scaling metadata (\texttt{fan\_in}, \texttt{fan\_out}, scaling type) to each parameter tensor, and each of these infrastructure layers either destroys or invalidates that metadata. Through our distributed u-\mup training wrapper, \texttt{dd\_unit\_scaling}, we address the following:

\paragraph{\texttt{torch.compile} compatibility.}
We rewrote the autograd scaling functions to eliminate graph breaks and cache distributed state before compilation.

\paragraph{FSDP2.}
FSDP2 replaces parameter tensors with DTensors, which destroys any attached metadata. We cache all \mup{} metadata by parameter name before sharding so it survives the replacement.

\paragraph{Data/Tensor parallelism.}
All batch-dependent scale factors are computed from the global effective batch: \texttt{local\_batch} $\times$ \texttt{world\_size} $\times$ \texttt{accumulation\_steps}. Loss is multiplied by \texttt{world\_size} to undo DDP's gradient averaging.

\paragraph{Sequence-length invariance.}
Unit-scaled attention has scale factors that depend on sequence length, which breaks KV caching (vital for production inference) since the effective length changes between decoding steps. We disable unit scaling in attention and the MLP activations. However, we still use the \mup{}-standard $1/\dk$ scale for scaled dot-product attention. The resulting variance mismatch between residual branches is mitigated by setting $\alpha_{\text{res-attn-ratio}} = \sqrt{S / \log S}$, where $S = \text{context\_length}/\text{patch\_size}$, and setting $\alpha_{\text{res}} = 0.75$.

We provide \texttt{dd\_unit\_scaling} to the community as an open-source, general-purpose library. We built it for \toto, but it is useful for anyone training under \ump{} at scale beyond what the upstream library was designed for.
\begin{figure}[!t]
    \centering
    \includegraphics[width=\linewidth]{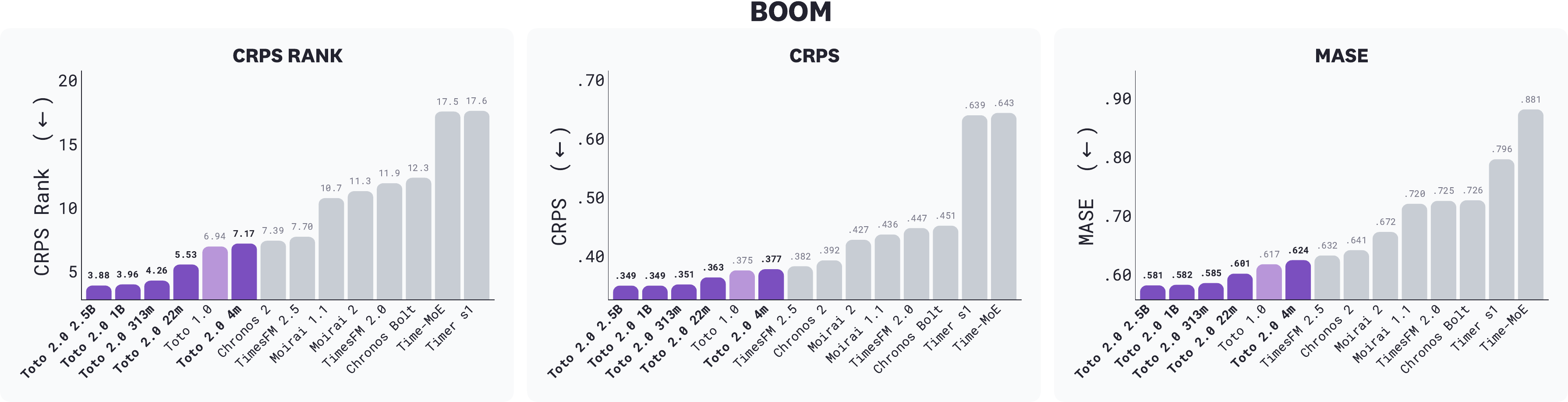}
    \caption{\boom{} results across CRPS rank, CRPS, and MASE; lower is better. All five \totov{} sizes outrank every other foundation model on every metric. \totov{} 22m matches or beats \totoone{} across all three with roughly $7\times$ fewer parameters. \totov{} models are shaded in purple.}
    \label{fig:bar_boom}
\end{figure}

\section{Results}
\label{sec:results}

We evaluate \totov{} on three forecasting benchmarks: \boom{}, our observability benchmark; \gifteval{}, the standard general-purpose benchmark; and TIME, a recent contamination-resistant zero-shot benchmark constructed from fresh datasets specifically chosen to mitigate the test-set contamination that affects established benchmarks. 

\textbf{\totov{} sets a new state of the art on all three.} Every \totov{} size leads on \boom{}. The three largest \totov{} sizes lead foundation models on \gifteval{}, and 2.5B-FT and \ff{} ensemble take the top two spots outright. On TIME, the same larger sizes take the top three spots on every metric, ahead of every external foundation model evaluated (\Cref{fig:bar_time}). 

Beyond accuracy, \Cref{sec:latency} examines inference latency, where every \totov{} size beats \totoone{} at long horizons, and \Cref{sec:long_horizon} probes long-horizon stability, showing how larger sizes retain coherent multi-scale structure well past their training context.

\paragraph{Benchmark setup.} All three benchmarks report results across several metrics. \emph{CRPS} (Continuous Ranked Probability Score) measures the quality of a probabilistic forecast, scoring how well a predicted distribution over future values aligns with observed outcomes; it is the metric most directly relevant to production forecasting use cases. \emph{MASE} (Mean Absolute Scaled Error) measures point forecast accuracy normalized against a naive seasonal baseline. Where metrics are reported as ranks, scores are averaged across all benchmark datasets to enable comparison across heterogeneous data. 

We use a context length of 2{,}048 on \boom{} and 4{,}096 on \gifteval{}; TIME prescribes a per-task context length aligned with each task's horizon, which we use as specified. Internal missing values in the context gaps are forward-filled, and the causal scaler's location and scale are backfilled on leading patches with fewer than 8 observations. At decode time, each real-space output quantile is clamped to the observed context's min and max, each extended by $10^{4}$ times the anchor scale at the final context position.

\begin{figure}[!t]
    \centering
    \includegraphics[width=\linewidth]{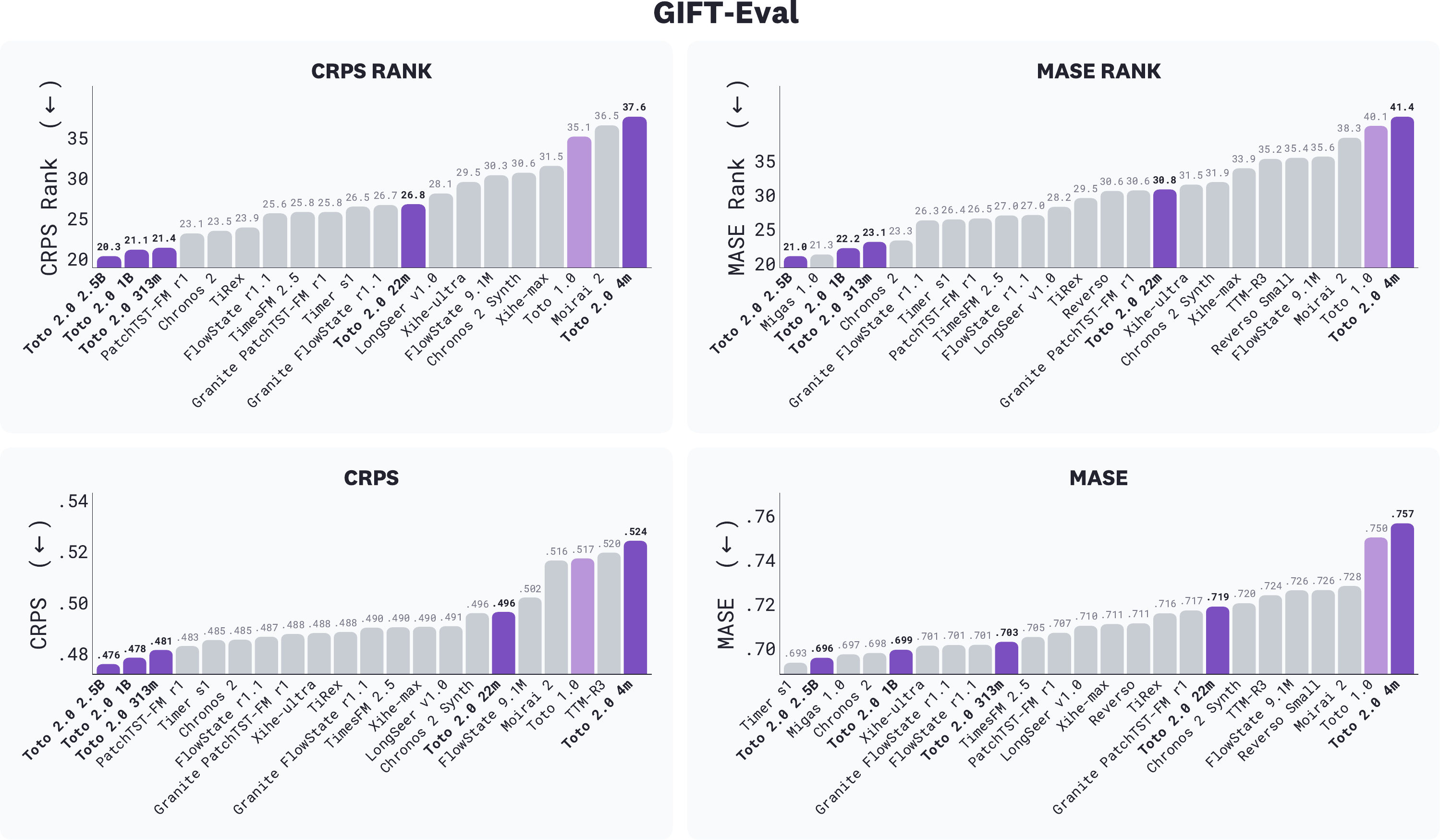}
    \caption{\gifteval{} results, filtered to foundation models only (i.e.,\ excluding finetuned, ensemble, and agentic systems), across CRPS rank, MASE rank, CRPS, and MASE; lower is better. \totov{} sizes are highlighted in purple. \totov{} sizes claim the top three spots on CRPS rank; the 2.5B alone leads on MASE rank.}
    \label{fig:bar_gifteval}
\end{figure}

\begin{figure}[!t]
    \centering
    \includegraphics[width=\linewidth]{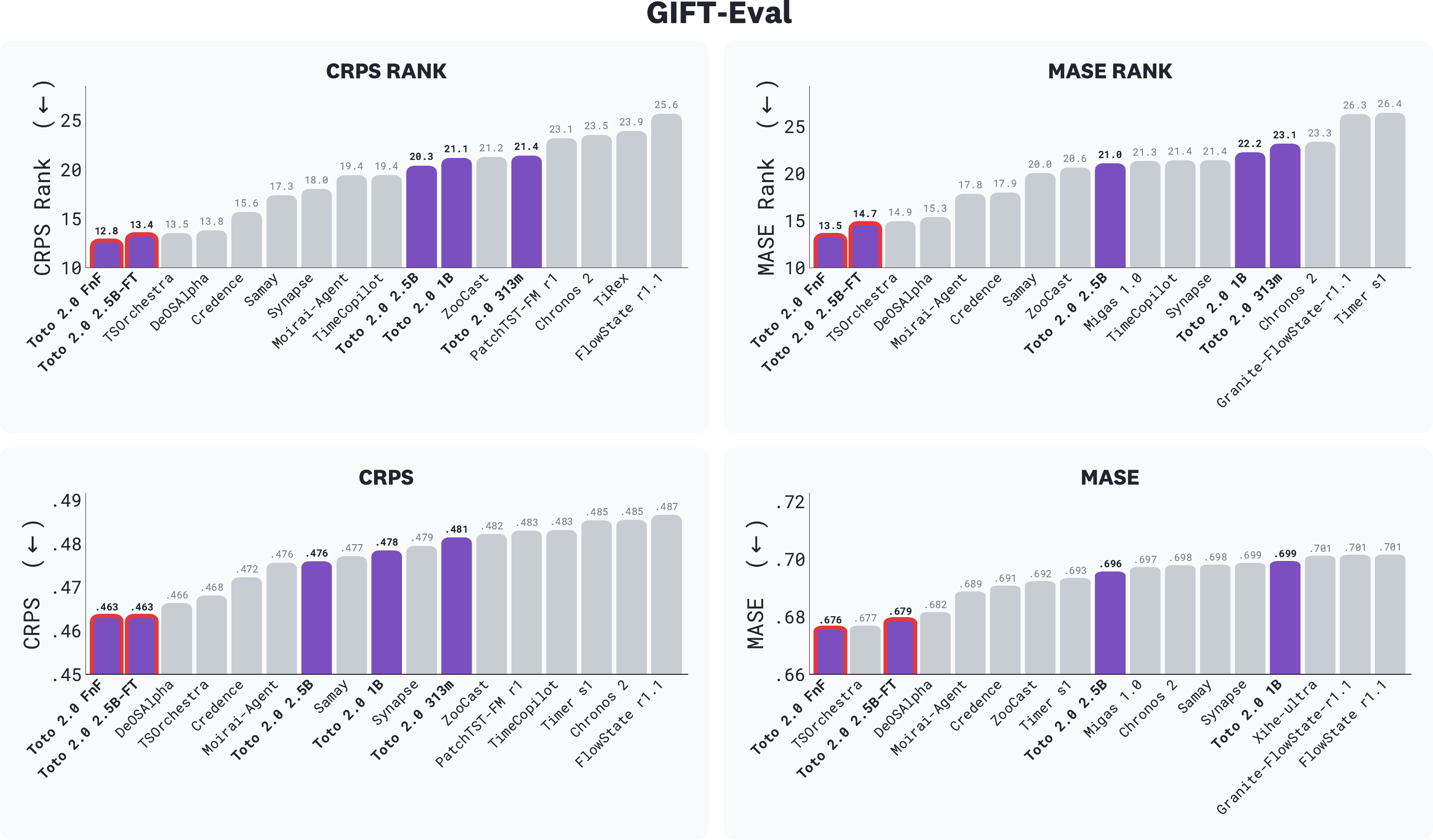}
    \caption{\gifteval{} leaderboard showing all submission types: foundation models, finetuned models, ensembles, and agentic systems together. On this leaderboard, \textquotedblleft finetuned\textquotedblright{} is used as an umbrella term for any model that uses the \gifteval{} training split, including ensemble and agentic systems. Our finetuned and ensemble models are highlighted in pink. The \ff{} ensemble ranks first on every metric (tied on raw CRPS), and the finetuned \totov{} 2.5B ranks second on the rank metrics and third on the raw metrics.}
    \label{fig:bar_gifteval_ft}
\end{figure}

\subsection{\boom}
\label{sec:boom}

\boom{} evaluates forecasting on observability metrics like CPU utilization, memory, request latency, and error rates. These are the signals production monitoring systems care about.

Every \totov{} size sits on the Pareto frontier of \boom{} (\Cref{fig:bar_boom}): at any given parameter count, no other foundation model produces better forecasts. The three largest sizes lead the chart with CRPS ranks of 3.88 (2.5B), 3.96 (1B), and 4.26 (313m). Behind them, the 22m at 5.53 already clears \totoone{} (6.94), establishing a $\sim 7\times$ parameter-efficiency improvement over \totoone{} (which has 151m parameters). The 4m, at 7.17, is competitive with \totoone{} and Chronos-2 (7.39) despite being $\sim 38\times$ smaller, making it a strong option for edge deployment.

\subsection{\gifteval{} -- foundation models}
\label{sec:gifteval}

\gifteval{} spans 97 evaluation tasks (combinations of dataset, frequency, and prediction horizon) drawn from 23 base datasets across domains like energy, retail, weather, and finance.

While most models train on a large collection of public domain data, \totov{} ranks first among foundation models on \gifteval{} (\Cref{fig:bar_gifteval}) despite training only on synthetic and observability data (\Cref{sec:data}). The three largest sizes score 20.3 (2.5B), 21.1 (1B), and 21.4 (313m) on CRPS rank, with a 1.7-point gap separating the 313m from the next best foundation model, PatchTST-FM~r1~\citep{nie2023patchtst} at 23.1. Chronos-2, a strong competitor, sits at 23.5. The 22m at 26.8 beats \totoone{} (35.1) by more than 8 points. On \gifteval{}, each successive \totov{} size improves over the one below it on the rank metrics.

\subsection{\gifteval{} -- finetuned and ensemble models}
\label{sec:gifteval_ft}

The results in this section are not used to support the zero-shot scaling claim; they show that the \totov{} base family is a strong starting point for downstream adaptation. The \gifteval{} leaderboard includes entries for finetuned foundation models (tuned on the benchmark's official training split), as well as agentic and ensembling methods that combine multiple foundation models. We explored both: finetuning a single model on a mix that includes the \gifteval{} train split (Toto 2.0 2.5B-FT), and ensembling multiple models with a learned per-window weighting scheme (Toto 2.0 FnF).

\paragraph{Finetuning.} \gifteval{} ships with two separate public datasets, both of which we use here: \gifteval{} \emph{Pretrain}~\citep{gifteval_pretrain_hf}, a large companion pretraining corpus curated to not overlap with the benchmark's evaluation datasets; and the official train splits of those evaluation datasets themselves~\citep{gifteval_hf}, which we refer to as \gifteval{} \emph{train}. Only the latter places a submission in the leaderboard's finetuned tier. We finetuned the 2.5B \totov{} base model for 10{,}000 steps from a fully-decayed base checkpoint on a mix of these two sources plus Datadog observability data. The full mix was: \gifteval{} Pretrain (45\%), Datadog 5+\,minute metrics (25\%), \gifteval{} train (15\%), synthetic (10\%), and Datadog 10\,s and 60\,s metrics (2.5\% each), with the \gifteval{} Pretrain portion drawn from the \totoone{} public-data pool of \gifteval{} Pretrain and the Chronos pretraining corpus~\citep{ansari2024chronos} (non-leaking). We also reduced the \normuon{} and \adamw{} learning rates by roughly an order of magnitude from pretraining, to 0.05 and 0.001, respectively.

\paragraph{Ensembling.} Forecasting datasets reward different model strengths: some favor strong short-horizon priors, others broad pretraining coverage \ff{} is an ensemble approach that picks per-window weights over a pool of ten foundation models: all five \totov{} sizes plus Chronos-2~\citep{ansari2025chronos2}, TimesFM~2.5~\citep{google2025timesfm}, TiRex~\citep{auer2025tirex}, FlowState~\citep{graf2025flowstate}, and PatchTST-FM~r1~\citep{nie2023patchtst}.

\ff{} follows the FFORMA (Feature-based FORecast Model Averaging) framework \citep{monteromanso2020fforma}, with an XGBoost regressor \citep{chen2016xgboost} as the meta-learner. The regressor consumes lightweight summary features extracted from each input window -- statistical moments, autocorrelation, seasonality, frequency, and horizon, extracted with the tsfeatures library \citep{garza2022tsfeatures} -- and emits softmax-normalized weights over the model pool. We train one head per (frequency, horizon-term) bucket, twenty in total, to handle \gifteval{}'s heterogeneity. We then adapt the overall weighted average (OWA) metric \citep{makridakis2020m4} for the GIFT-Eval leaderboard. For a model $f$ in the candidate pool, and window \(j\) of a dataset, the OWA is defined as 

\[
    \mathrm{OWA}_{f,j} = \frac{1}{2}\left(\frac{\mathrm{MASE}_{f,j}}{\mathrm{MASE}_{\mathrm{sNaive}}} + \frac{\mathrm{CRPS}_{f,j}}{\mathrm{CRPS}_{\mathrm{sNaive}}}\right)
\]

where \(\mathrm{MASE}_{\mathrm{sNaive}}\) and \(\mathrm{CRPS}_{\mathrm{sNaive}}\) are computed from the seasonal naive baseline, across train windows in the dataset.

\textbf{Both place at the top of the GIFT-Eval leaderboard} (\Cref{fig:bar_gifteval_ft}): \ff{} ranks first on every metric (tied with TSOrchestra on raw CRPS), and the finetuned 2.5B ranks second on the rank metrics and third on the raw metrics. 

But the more interesting finding is what is inside the ensemble. The meta-learner's softmax weights reveal what each candidate actually contributes to each prediction. Averaged across all predictions, the \totov{} family accounts for 39\% of the assigned weight, more than any other model in the pool, ahead of Chronos-2 (32\%) and more than the four remaining external models combined. The ensemble does not replace \totov{}; instead it confirms that, when the meta-learner is free to weight everything available to it, the learner consistently spends more on the \totov{} family than on any other source.

\begin{figure}[!ht]
    \centering
    \includegraphics[width=\linewidth]{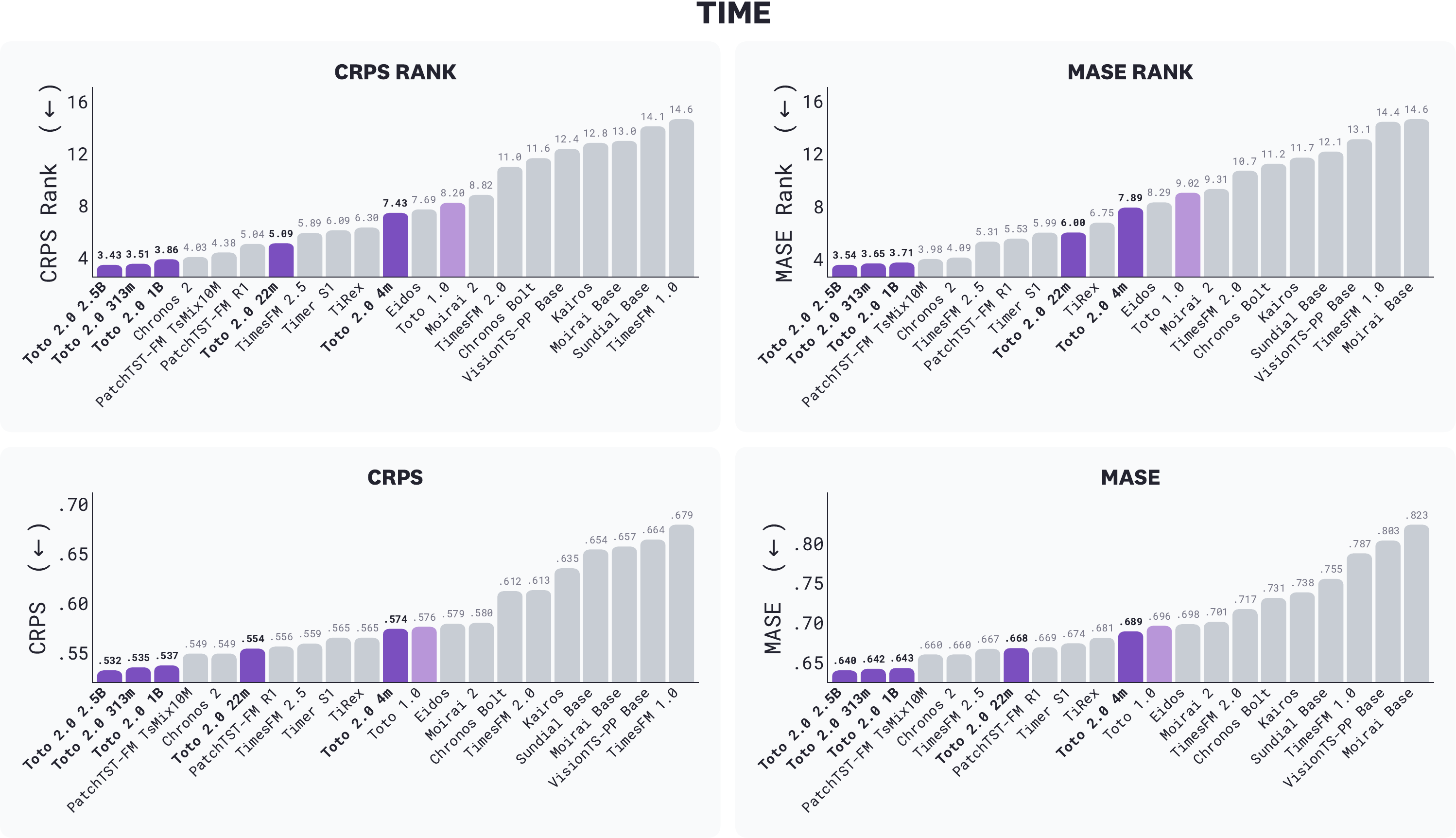}
    \caption{Results on TIME across CRPS rank, MASE rank, CRPS, and MASE; lower is better. \totov{} sizes are highlighted in purple. \totov{} sizes take the top three slots on every metric. The 2.5B leads on CRPS rank, MASE rank, and CRPS; the 313m leads on MASE and edges out the 1B on the two rank metrics, the one place the family departs from the otherwise clear size-vs.-quality trend.}
    \label{fig:bar_time}
\end{figure}

\subsection{TIME}
\label{sec:time}

We additionally evaluate on TIME~\citep{qiao2026time}, comprising 98 forecasting tasks drawn from 50 ``fresh'' (never/rarely been explored by existing TSF benchmarks) datasets curated under a human-in-the-loop pipeline, with horizons aligned to real-world operational requirements rather than mechanical short/medium/long buckets. The benchmark deliberately avoids legacy datasets such as ETTh1, Electricity, Traffic, and Weather that have circulated through TSFM pretraining corpora for years, replacing them with recent data unlikely to have been seen during pretraining.

\textbf{\totov{} takes the top three spots on every TIME metric} (\Cref{fig:bar_time}). The 2.5B leads on CRPS rank (3.43), MASE rank (3.54), and CRPS (0.532). The strongest external foundation models, Chronos-2~\citep{ansari2025chronos2} and PatchTST-FM~r1~\citep{nie2023patchtst}, trail the \totov{} top three on every metric, with Chronos-2 fourth on CRPS rank (4.03) and PatchTST-FM~r1 fifth (5.04). Scaling on TIME is not strictly monotonic within the \totov{} family: the 313m leads on MASE and edges out the 1B on both rank metrics---the only point at which the family departs from a clear size-vs.-quality trend (\Cref{fig:bar_time}). Every \totov{} size, including the 4m, still outperforms \totoone{}. 

\begin{figure}[!t]
    \centering
    \includegraphics[width=\linewidth]{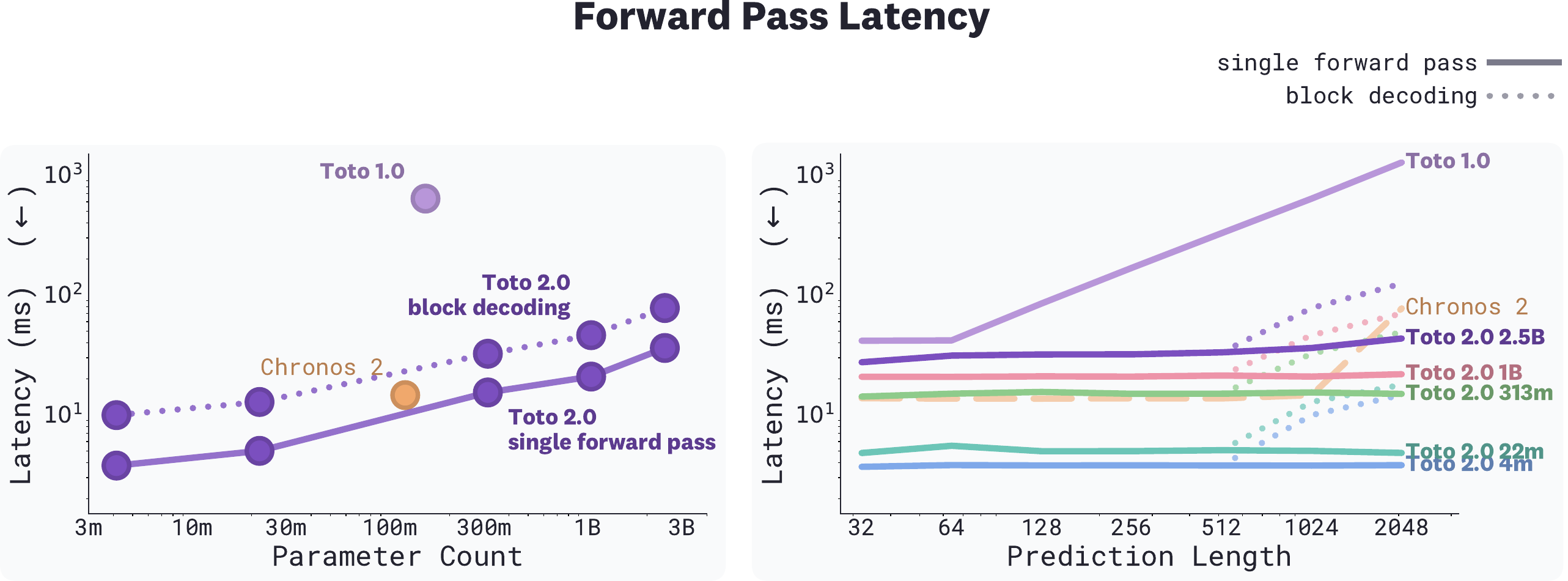}
    \caption{\textbf{Left}: forward pass latency vs.\ parameter count at forecast length$=$1{,}024. Every \totov{} size is significantly faster than \totoone{}. \textbf{Right}: forward pass latency vs.\ forecast horizon (log scale). \totov{} stays flat in single-pass mode up to a 768-point forecast length, which we found best on synthetic signals. At a forecast horizon of 4{,}096 steps, 2.5B in single-pass mode remains faster than Chronos-2.}
    \label{fig:latency}
\end{figure}

\subsection{Inference latency}
\label{sec:latency}

CPM does not just improve forecast quality; it makes \totov{} dramatically faster. The two decoding modes, single-pass and block decoding (\Cref{sec:cpm}), trade off speed for long-horizon stability. Single-pass runs the entire horizon in one forward pass and is what we use for the leaderboard submissions above. Block decoding generates the horizon in segments, conditioning each on the previous segment's median, with KV caching for efficiency.

We evaluate forward pass latency against \totoone{} and Chronos-2, the previous state of the art on \gifteval{}. A 1{,}024-step forecast takes \totoone{} up to 16 autoregressive steps and single-pass \totov{} a single forward pass. Every \totov{} size is significantly faster than \totoone{} at this horizon, and the 313m runs at roughly the same latency as Chronos-2 (120m parameters)~\citep{ansari2025chronos2} (\Cref{fig:latency}).

\begin{figure}[!t]
    \centering
    \includegraphics[width=\linewidth]{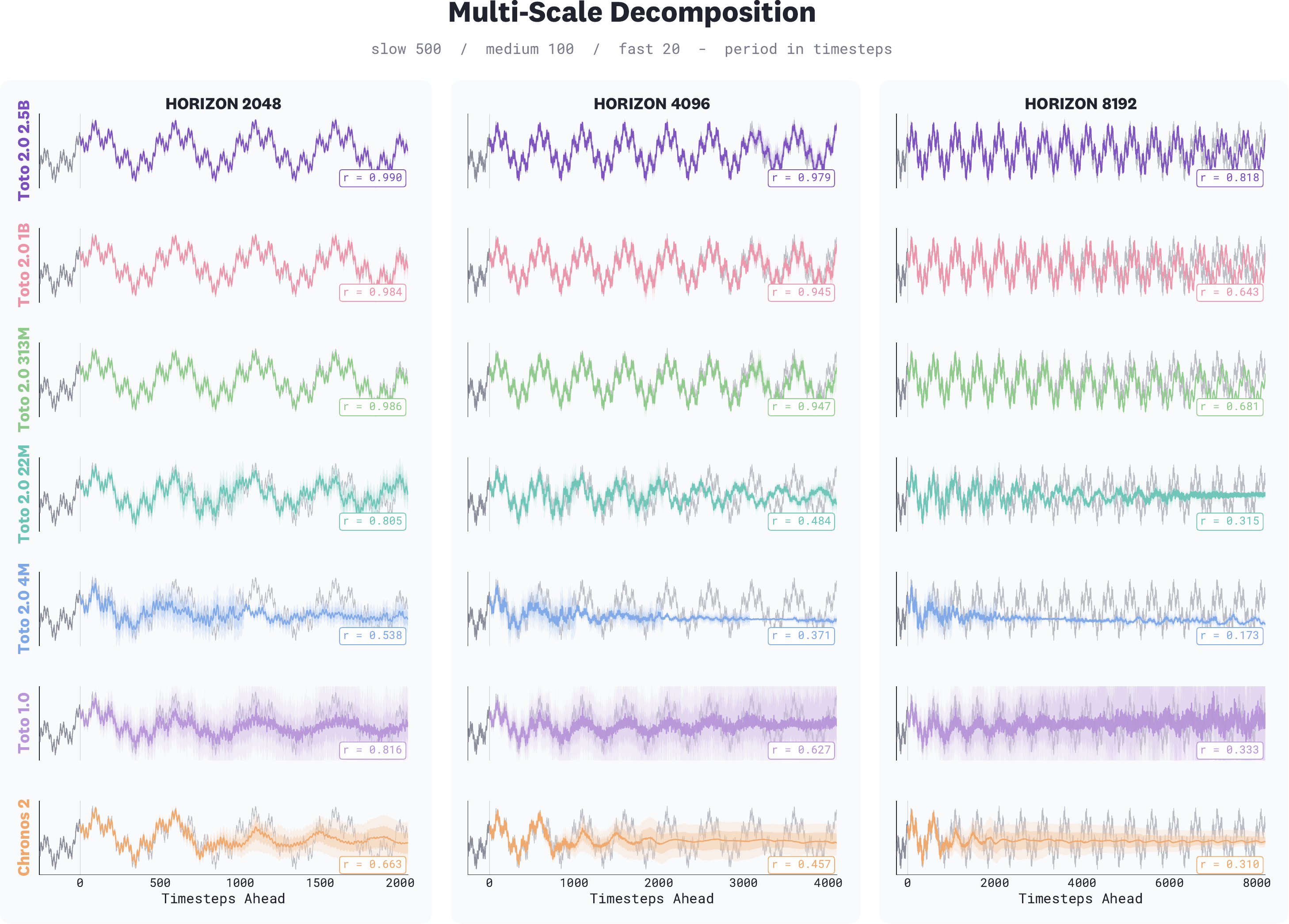}
    \caption{Forecasts on a synthetic multi-scale signal (superimposed periods of 500, 100, and 20 timesteps) at three forecast horizons (2{,}048, 4{,}096, and 8{,}192 steps). Each row is a model, each column a horizon. Ground truth is plotted in gray; model forecasts in color. Larger \totov{} sizes maintain coherent multi-scale structure at 8{,}192 steps; smaller sizes and prior-generation models lose structure progressively. Pearson correlation against ground truth is shown in each panel. \totov{} forecasts use block decoding (\Cref{sec:cpm}). This experiment is illustrative: it measures stability beyond the training horizon on synthetic signals, not extrapolation to genuinely novel dynamics.}
    \label{fig:long_horizon}
\end{figure}

\subsection{Long-horizon stability}
\label{sec:long_horizon}

Forecast quality on benchmarks like \boom{} and \gifteval{} reflects how a model performs within or near its training context.  But many practical tasks want both long horizons and fine resolution. Downsampling buys horizon at the cost of the high-frequency structure (e.g. spikes, transient anomalies, sub-period dynamics, etc.) that the forecast is meant to capture. 

To understand how \totov{} behaves when asked to forecast much further than it was trained on, we evaluated all five sizes on randomly-generated sinusoidal mixtures at horizons of 2{,}048, 4{,}096, and 8{,}192 timesteps (well past the 4{,}096-step training context used for \toto). This is an illustrative stability test: it measures behavior beyond the training horizon, not extrapolation to genuinely novel dynamics.

We compare all five \totov{} sizes to \totoone{} and Chronos-2 across the three horizons (\Cref{fig:long_horizon}). 4m captures short-range patterns but collapses past its training context, producing flat or noisy forecasts. 22m holds longer but degrades by a 4{,}096-step forecast horizon. 313m is stable through 4{,}096 but loses structure beyond. 1B maintains the underlying pattern across all three horizons; 2.5B is more accurate still. \totoone{} and Chronos-2, despite Chronos-2 being trained on longer sequences, both lose coherence well before the 1B does.
\section{Discussion}
\label{sec:discussion}

\totov{} is the first TSFM family for which simply making the model bigger reliably makes it better. A single recipe applied across widths produces smooth improvements on \boom{}, \gifteval{}, and TIME from 4m up to 2.5B parameters, with only minor inversions inside the family on TIME's rank metrics (\Cref{sec:time}). If \totoone{} and its contemporaries were the field's BERT~\citep{devlin2019bert} moment~\citep{berts2025workshop}, \totov{} is similar in some respects to a GPT-2 moment~\citep{radford2019gpt2}: scaling TSFMs is no longer a research question but a tool. Continuing to scale---more data, larger models---is a natural direction for future work. Below we outline what we see as the other major open questions for TSFM research:

\paragraph{Closing the gap with classical baselines.}
Foundation models capture dynamics classical statistical methods largely cannot: multivariate interactions, long context, and transfer across domains. But classical methods still have properties foundation models lack: clean extrapolation on simple signals, appropriate growth of prediction intervals with horizon under well-specified models, and predictable behavior on out-of-distribution samples. The long-horizon study in \Cref{sec:long_horizon} (\Cref{fig:long_horizon}) is one window into this. Even the 2.5B loses some structure at a forecast horizon of 8{,}192 steps where a properly-fitted seasonal model would extrapolate cleanly. The gap shows up in many places: tail behavior, regime shifts, and forecasts on signals far outside any plausible training context. Closing it will likely require several things in combination: targeted architectural changes, continued scaling, and novel post-training objectives.

\paragraph{Improved data curation.}
Data curation in TSFMs has been ad hoc. Models typically mix synthetic series and a few public (or private) datasets, sample frequencies in proportions chosen by hand or by sweep, and stop there. In language modeling, data curation is treated as a first-class research problem: quality filtering, deduplication, annotation, mixing, curriculum. TSFM research has not gotten there yet, partly because scaling itself was still the open question: curation is a luxury you can only afford once data is abundant. In our own hyperparameter sweep, the optimal mix for pretraining excluded public data entirely (\Cref{sec:hp_round2}), while the optimal mix for finetuning was 45\% public (\Cref{sec:gifteval_ft}). These are not intuitive results, and we arrived at them empirically rather than through principled selection. With scaling now reliable, it is time to take curation more seriously.

\paragraph{Metrics as a distinct modality.}
With Toto 1.0 and 2.0, we have built TSFMs suited for generic time series found commonly in the open. However, here at Datadog, we are interested in modeling the massive amounts of metrics data\footnote{\url{https://docs.datadoghq.com/metrics/}} that we collect. While we have been able to cast Datadog metrics as basic time series, they are in fact a distinct data modality with unique properties. By compressing them into the mold of generic time series data, we lose  significant amounts of embedded information and structure. In future work, we aim to prioritize the unique challenges of modeling Datadog metrics. 
Firstly, our architecture should be able to cater to the various metric types found on the Datadog platform, including histogram and distribution type data. Secondly, we deal with real world time series which have complex seasonality, such as multiple seasonality across long contexts, as well as non-integer and uneven periods. Thirdly, our data contains complex multivariate structure, including heterogeneous frequency where multivariate series can be sampled at different frequencies, as well as a context selection problem, where we have extremely high-dimensional series and we face the problem of selecting the relevant variates for the task at hand.

\paragraph{Multimodality and world models for observability.}
While multimodality for time series models has become an increasingly hot topic, it predominantly focuses on time series + text with limited datasets and evaluations \citep{liu2024timemmd,liu2026rethinking,xu2025fidelts,chang2025timeimm}. At Datadog, we care about models that understand how distributed systems behave. Our observability data is diverse and comprehensive, meaning we can develop models that deal not just with metrics, but also traces, logs, topology, code changes, events, alerts, text, etc. 
Our first step in this direction has been our recently released ARFBench~\citep{xie2026arfbench}, which focused on evaluating incident-grounded multimodal reasoning. 
Our longer-term goal is to develop a full-fledged world model for observability, extending to all telemetry types, unlocking capabilities such as proactive incident detection, root cause analysis, counterfactual analysis, simulation, and agent training.

\section*{Acknowledgements}
We thank Clement Acher, Askar Aitzan, Taha Aksu, Bogna Blaszczyck, Etienne Brodu, Ben Cohen, Antonin Couturier, Walid Elbouchikhi, Quentin Gendre Robin, Howard Huang, Sarra Kazdaghli, Mikhail Khodak, Shridhar Kumar, Rohan Kulkarni, Salahidine Lemaachi, Gael Magnan, Savita Manghnani, Hugo Miccinilli, Samuel Mueller, Ali Naeimi, Matthieu Neau, Sergey Pastukhov, Qiqi Ren, Afshin Rostamizadeh, Anna-Monica Toon, Lucas Verdonk, Kan Wang, and Stephan Xie for valuable discussions, infrastructure support, and contributions to the broader Toto effort.

\bibliographystyle{plainnat}
\bibliography{refs}

\end{document}